\documentclass[sigconf,nonacm=true]{acmart}

\AtBeginDocument{%
  \providecommand\BibTeX{{%
    \normalfont B\kern-0.5em{\scshape i\kern-0.25em b}\kern-0.8em\TeX}}}

\setcopyright{acmcopyright}
\copyrightyear{2022}
\acmYear{2022}
\acmDOI{TBD} %TODO

\acmConference[GECCO ’22]{Woodstock '18: ACM Symposium on Neural
  Gaze Detection}{July 09–-13, 2022}{Boston, USA}
\acmBooktitle{GECCO ’22: Genetic and Evolutionary Computation Conference,
  July 09–-13, 2022, Boston, USA}
\acmPrice{} %TODO
\acmISBN{TBD} %TODO

\usepackage{libertine}
\usepackage{siunitx}
\usepackage{multirow}
\usepackage{xcolor,colortbl}
\usepackage{hhline}
\usepackage{nicematrix}
\usepackage{array}
\usepackage{graphicx}
\usepackage[export]{adjustbox}
\usepackage{enumitem}
\usepackage{setspace}

\usepackage[ruled,noend]{algorithm2e}

\makeatletter
\def\maketag@@@#1{\hbox{\m@th\normalfont\normalsize#1}}
\makeatother

\begin{document}

%%
%% The "title" command has an optional parameter,
%% allowing the author to define a "short title" to be used in page headers.
\title
[Adaptive Objective Configuration in Bi-Objective Evolutionary Optimization] %short
{Adaptive Objective Configuration in Bi-Objective Evolutionary Optimization for Cervical Cancer Brachytherapy Treatment Planning} %full

%%
%% The "author" command and its associated commands are used to define
%% the authors and their affiliations.
%% Of note is the shared affiliation of the first two authors, and the
%% "authornote" and "authornotemark" commands
%% used to denote shared contribution to the research.

\author{Leah R.M. Dickhoff}
\orcid{https://orcid.org/0000-0001-6720-4380}
\affiliation{%
 \institution{Leiden University Medical Center}
 \city{Leiden}
 \country{The Netherlands}}
\email{L.R.M.Dickhoff@lumc.nl}

\author{Ellen M. Kerkhof}
\orcid{https://orcid.org/0000-0002-6070-7732}
\affiliation{%
 \institution{Leiden University Medical Center}
 \city{Leiden}
 \country{The Netherlands}}
\email{E.M.Kerkhof@lumc.nl}

\author{Heloisa H. Deuzeman}
\orcid{https://orcid.org/0000-0003-3282-8497}
\affiliation{%
 \institution{Leiden University Medical Center}
 \city{Leiden}
 \country{The Netherlands}}
\email{H.H.Deuzeman@lumc.nl}

\author{Carien L. Creutzberg}
\orcid{https://orcid.org/0000-0002-7008-4321}
\affiliation{%
 \institution{Leiden University Medical Center}
 \city{Leiden}
 \country{The Netherlands}}
\email{C.L.Creutzberg@lumc.nl}

\author{Tanja Alderliesten}
\orcid{https://orcid.org/0000-0003-4261-7511}
\affiliation{%
 \institution{Leiden University Medical Center}
 \city{Leiden}
 \country{The Netherlands}}
\email{T.Alderliesten@lumc.nl}

\author{Peter A.N. Bosman}
\affiliation{%
 \institution{Centrum Wiskunde \& Informatica}
 \city{Amsterdam}
 \country{The Netherlands}}
\email{Peter.Bosman@cwi.nl}

%%
%% By default, the full list of authors will be used in the page
%% headers. Often, this list is too long, and will overlap
%% other information printed in the page headers. This command allows
%% the author to define a more concise list
%% of authors' names for this purpose.

\renewcommand{\shortauthors}{Dickhoff, et al.}

%-----------------------------------------------------------------------------
%%
%% The abstract is a short summary of the work to be presented in the
%% article.
\begin{abstract}   % 198 /200 words

The Multi-Objective Real-Valued Gene-pool Optimal Mixing Evolutionary Algorithm (MO-RV-GOMEA) has been proven effective and efficient in solving real-world problems. A prime example is optimizing treatment plans for prostate cancer brachytherapy, an internal form of radiation treatment, for which equally important clinical aims from a base protocol are grouped into two objectives and bi-objectively optimized. This use of MO-RV-GOMEA was recently successfully introduced into clinical practice. Brachytherapy can also play an important role in treating cervical cancer. However, using the same approach to optimize treatment plans often does not immediately lead to clinically desirable results. Concordantly, medical experts indicate that they use additional aims beyond the cervix base protocol. Moreover, these aims have different priorities and can be patient-specifically adjusted. For this reason, we propose a novel adaptive objective configuration method to use with MO-RV-GOMEA so that we can accommodate additional aims of this nature. Based on results using only the base protocol, in consultation with medical experts, we configured key additional aims. We show how, for 10 patient cases, the new approach achieves the intended result, properly taking into account the additional aims. Consequently, plans resulting from the new approach are preferred by medical specialists in 8/10 cases.

\end{abstract}

%%
%% The code below is generated by the tool at http://dl.acm.org/ccs.cfm.
%% Please copy and paste the code instead of the example below.
%%


%%
%% Keywords. The author(s) should pick words that accurately describe
%% the work being presented. Separate the keywords with commas.

%%
%% This command processes the author and affiliation and title
%% information and builds the first part of the formatted document.
\maketitle

%-----------------------------------------------------------------------------

\section{Introduction}

\setlength{\parskip}{.2pt plus .2pt}

Brachytherapy (BT) is a form of internal radiation treatment during which a radioactive source is guided to locations in and around the tumor to be treated. While radiation dose which is given off by this radioactive source kills nearby cancer cells, it may also harm nearby normal cells.
%harms nearby cells. 
Hence, the goal is to deliver sufficient dose to the cancer cells to eliminate them, 
%enough of this dose to the cancerous cells so that they die, 
while dose to surrounding healthy Organs At Risk (OARs) should be limited as much as possible. For cervical cancer, the guidance of the radioactive source to the tumor is achieved via an applicator, which is placed inside the uterus and the cervix, possibly accompanied by needles (catheters) applied through the vaginal mucosa. An example of such a setup is visualized in the Supplementary Material Figure 1. \par
Inside this applicator and catheters, the source can reside at different locations, called dwell positions, for different times, called dwell times. The longer the source resides at a dwell position, the more radiation dose is given off to the subvolume around it. As such, a treatment plan is defined by a set of dwell times, ideally achieving the desired patient-specific trade-off between coverage of the region that should be irradiated (i.e., target volume) and sparing of the surrounding OARs.\par
As there are numerous dwell times to configure (for cervical BT, typically, around 80) and treatment planning needs to be done within a limited time window, automation is highly desirable and needed to find the best possible treatment plans for a specific patient. This has been successfully accomplished by an automatic treatment planning approach in BT for prostate cancer using a modern evolutionary algorithm \cite{Luong2018ApplicationTreatment} \cite{Bouter2019GPU-acceleratedBrachytherapy}. In this approach, the coverage-sparing trade-off is intuitively captured through a bi-objective formulation. The clinical aims in terms of planned dose to each of the target volumes and OARs are laid out in a clinical base protocol, and directly included as aspiration values in this model. The approach uses a tailored version of the Multi-Objective Real-Valued Gene-pool Optimal Mixing Evolutionary Algorithm (MO-RV-GOMEA), implemented on a GPU requiring only a few minutes  runtime \cite{Bouter2019GPU-acceleratedBrachytherapy}. The output is a set of treatment plans, each of which is characterized by a different high quality trade-off between coverage and sparing. This provides insight into what can be achieved for each specific patient, and allows the physician to intuitively choose the desired treatment plan. This (semi-)automatic treatment planning approach has been incorporated into clinical practice at a medical center.\par
Besides prostate cancer, BT was proven highly effective and essential for cure of cervical cancer \cite{Chargari2019Brachytherapy:Clinicians}, and for which different optimization techniques have been developed \cite{Trnkova2010ABrachytherapy} \cite{Oud2020}. Still, automatic treatment planning is in this case scarcely used in practice. Existing methods do not use the bi-objective approach however, and often smooth surrogate functions of the actual clinical aims, which is not needed for the implementation based on MO-RV-GOMEA. We therefore want to extend the MO-RV-GOMEA-based bi-objective approach to cervical cancer. The EMBRACE II protocol \cite{EMBRACE} is for this case the officially recommended base protocol to follow. However, previous research has shown that the treatment planning aims specified in EMBRACE II are not sufficient to automatically generate a set of clinically acceptable treatment plans, as medical experts are found to use additional criteria in practice \cite{ESTROblinded}.\par
Hence, to still be able to automate the generation of high quality clinically acceptable treatment plans, proper consideration of an extra set of aims is needed, on top of the aims of EMBRACE II. This set of added aims differs from those of EMBRACE II in three ways. 1) The added aims can be institution-specific whereas EMBRACE II is considered a standard protocol in which the aspiration values are based on clinical evidence; hence, the added aims are less important than those of EMBRACE II. 2) Organ locations, tumor extent, and dwell position geometry are vastly patient-specific, implying that, for each patient, different aspiration values may be needed for each added aim. 3) Not all of the added aims are of equal priority.\par
Since the MO-RV-GOMEA implementation for prostate BT only optimizes on a base protocol with equally important aims, the resulting plans can merely satisfy this one set of aims. This was deemed not clinically acceptable, since many manual (patient-specific) plan adjustments would be needed, meaning that there would be virtually no gain with respect to manual planning. Therefore, in order to successfully bring the strengths of EA-based automated treatment planning from prostate BT to cervical BT, a new, augmented EA-based procedure is needed. Here, we introduce a way to still optimize the dwell times, but also, adaptively adjust the the aspiration values of the added aims during optimization. Hence, our new method makes use of MO-RV-GOMEA, but can include two different types of aims, which are of distinct importance, and of which one type comprises adaptable aims of different priorities.

\vspace*{-2pt}
\section{MO-RV-GOMEA} \label{sec:GOMEA}

GOMEA \cite{Thierens2011OptimalAlgorithms}, and its extension to problems with multiple objectives and real-valued variables MO-RV-GOMEA \cite{Bouter2021AchievingEvaluations},
%\cite{Bouter2017TheAlgorithm}, 
have been demonstrated to be efficient when optimizing both benchmark 
%\cite{Bouter2017ExploitingAlgorithm} 
as well as real-world application problems. 
%\cite{Bouter2017AQuality}.
This is especially the case when an optimization function permits partial evaluations, which can effectively be exploited by the algorithm.
%; this is for BT specifically further explained in Section \ref{sec:AdaptiveObjConfig}. 
For prostate BT, MO-RV-GOMEA has been shown to outperform multiple other multi-objective evolutionary algorithms and result in better solutions than previously used (mostly) manual optimization \cite{Luong2018ApplicationTreatment} \cite{Maree2019EvaluationStudy}.\par
As any EA, GOMEA maintains a population of individuals, or solutions to the problem at hand, which undergo selection and variation to create new solutions. Throughout a run, the non-dominated solutions are kept track of in an elitist archive of a fixed capacity (for which we used 1000). The following steps are performed each generation. First, the best (in terms of non-domination) solutions in the current population are selected. Then, this selection is clustered into multiple (we used 5) clusters of equal sizes. Both the size of this subset (here 35\% of the population size of 96) as well as the number of clusters are user defined settings \cite{Bouter2019GPU-acceleratedBrachytherapy}. Next, a linkage model is learned for each cluster separately by identifying subsets of variables which exhibit some degree of dependency, and should therefore be processed together during variation to prevent the disruption of important building blocks. Specifically in MO-RV-GOMEA, dependencies between variables are efficiently exploited by estimating a factorized normal probability distribution based on the sets of dependent variables (i.e., linkage sets) as identified by the linkage model. These sets can be determined online or offline - the latter being preferred if problem-specific information is available, since the former can cause overhead due to a need for a large population size. In the BT case, the offline linkage tree is learned from
the Euclidean distance between dwell positions. Finally, variations to the solutions are performed based on the estimated distributions: for each linkage set, new values are drawn from the distribution and tested inside the solution. If the solution improves, the change is kept, otherwise it is reversed. This is called optimal mixing.\par
Partial evaluations comprise the main way in which GOMEA's performance can significantly be increased. Often, when only a few variables were changed, the new objectives can be computed in notably less time than when modifications in all variables occurred, which is also the case in BT \cite{Bouter2019GPU-acceleratedBrachytherapy}.

\vspace*{-1pt}
\section{Cervical cancer Brachytherapy} \label{sec:brachyCervix}

\setlength{\parskip}{.2pt plus .2pt}

BT procedures can be of intracavitary and/or interstitial nature. Cervical cancer BT can be both, meaning that an intracavitary applicator is always placed within the natural body cavity of the uterus, with its base in the cervix, whereas additional interstitial needles (catheters) can be placed through the vaginal mucosa in order to reach more advanced tumors. %\par
Two ovoids are positioned in the vagina against the cervix, via which additional interstitial needles can be placed in the parametrial tissues.\par
Once the applicator and catheters are in place, medical images (by magnetic resonance imaging and/or computed tomography) are acquired, on which the target volumes and OARs are delineated. Cervical cancer BT includes three main target volumes - the High Risk Clinical Target Volume ($\mathrm{CTV_{HR}}$), the Intermediate Risk Clinical Target Volume ($\mathrm{CTV_{IR}}$), and the Residual Gross Tumor Volume ($\mathrm{GTV_{RES}}$) - depending on the spread of the cancer and risk of recurrence. OARs include the bladder, rectum, sigmoid, and bowel. The delineated structures can be seen for three patient cases in Figure \ref{fig:distributions}. Alongside these target volumes and OARs, dwell positions are also identified on the mentioned medical image.\par
Treatment planning concerns setting the dwell times for each of the dwell positions so that the planned radiation dose covers the target volumes but spares the OARs. In practice, this is most often still done manually, especially for cervical cancer. Medical specialists start with a standard setting for the dwell times in the applicator, then fine-tune all dwell times by hand.\par
One treatment delivery is called one fraction. Cervix High Dose-Rate (HDR) BT often includes 3-4 fractions, depending on local clinical practice. The time between subsequent fractions ranges from six hours to one week, subject to the number of fractions per applicator placement. Most patients at our university medical center, and all patients used for this study, were treated with 4 fractions of 7 \si{\gray} HDR each. The same applicator and catheter configuration was used for the first two (and the last two) fractions. Each patient case in this paper corresponds to a new applicator placement. A typical workflow of one BT treatment including the four fractions is presented in the Supplementary Material Figure 2.

\vspace*{1pt}
\section{Adaptive objective configuration} \label{sec:AdaptiveObjConfig}
In this section we describe how our approach for cervical cancer BT differs from the optimization previously used for prostate BT.

\vspace*{1pt}
\subsection{Treatment plan evaluation}

In clinical practice, treatment plan quality evaluation consists of both a visual inspection of the 3D dose distribution and the associated Dose Volume Index (DVI) values. During bi-objective optimization, only the latter are used.
Cervix BT can include three types of DVIs: a volume index $V_d^o$ denotes the subvolume of the Region Of Interest (ROI) $o$ which gets a dose of at least $d$, a dose index $D_v^o$ describes the minimum dose received by the most irradiated subvolume $v$ of $o$, and a dose point $D_{\text{point}}^{rp}$ defines the dose given to a specific reference point $rp$. All doses are given in percentages of the planned dose, which, for the patients used in this paper, is 7 \si{\gray}. 
Aspiration values for all DVIs are laid out in a clinical protocol; the DVI aims of EMBRACE II are presented in Table \ref{tab:EMBRACEprot}.\par
In our approach, the DVI values are approximated using Dose Calculation (DC) points sampled uniformly at random within each of the ROIs \cite{Niemierko1990RandomPlans}. The dose in these points is calculated using the TG-43 formalism that describes how much dose is delivered to a point at a certain distance from the source \cite{Rivard2004UpdateCalculations}. The dose values for each of these points can then be used to approximate, e.g., volumes with a certain property (such as all tissue receiving at most 2 \si{\gray}). 
Using large numbers of DC points, DVIs can be accurately computed. However, this can become prohibitively time consuming when used during optimization. Hence, MO-RV-GOMEA optimizes on a smaller amount of DC points \cite{Bouter2019GPU-acceleratedBrachytherapy}. Then, after optimization and prior to the selection of the preferred solution, the solutions are reevaluated on a larger amount of DC points in order to obtain accurate results. This implies that, when having optimized on too small a number of DC points, solutions can during reevaluation undergo a fallback in terms of their (perceived) 
objective values. 

\begin{table}[h]
    \small
    \centering
    \setlength\tabcolsep{2.4pt} %default: 6pt
    \renewcommand{\arraystretch}{1} %default: 1
%    \vspace*{-3mm}
    \begin{tabular}{|c|c|c|}
%         \toprule
         \multicolumn{3}{c}{\textbf{Target volumes}} \\\hline
         $\mathrm{CTV_{HR}}$ & $\mathrm{GTV_{RES}}$ & $\mathrm{CTV_{IR}}$  \\\hline
         \renewcommand{\arraystretch}{0.8}
           \begin{tabular}{@{}c@{}}
             $D_{90\%}>111\%$;\ \ $D_{90\%}<119\%$;\ \ $D_{98\%}>83\%$
           \end{tabular}\renewcommand{\arraystretch}{1} & $D_{98\%}>119\%$ & $D_{98\%}>50\%$  \\\hline
%         \bottomrule
    \end{tabular}
    \begin{tabular}{|c|c|c|c|c|}
%         \toprule
         \multicolumn{5}{c}{\textbf{OARs}} \\\hline
         Bladder & Rectum &  \renewcommand{\arraystretch}{0.8}\begin{tabular}{@{}c@{}}\ \\[-2.5mm]ICRU\\[-.9mm]rectovaginal\end{tabular}\renewcommand{\arraystretch}{1} & Sigmoid & Bowel \\\hline
         $D_{2\si{\cubic\centi\metre}}<78\%$ & $D_{2\si{\cubic\centi\metre}}<56\%$ & $D_{\mathrm{point}}<56\%$ & $D_{2\si{\cubic\centi\metre}}<64\%$ & $D_{2\si{\cubic\centi\metre}}<64\%$ \\\hline
%         \bottomrule
    \end{tabular}
    \caption{EMBRACE II protocol. A priority of 1 is attributed to all DVIs. Aspiration values are in percentages of 7 \si{\gray} physical dose. An ICRU point is recommended by the International Commission on Radiation Units and Measurements.}
    \label{tab:EMBRACEprot}
    \vspace*{-9mm}
\end{table}

\vspace*{-1pt}
\subsection{Added aims}
We added aims to EMBRACE II to tackle missing properties in the plans that were optimized on EMBRACE II only, as judged by medical experts. Table \ref{tab:addedprot} shows the DVIs, their possible dose aim ranges, and their priorities within the protocol, again as determined together with medical specialists. Note that all DVIs in the EMBRACE II protocol (Table \ref{tab:EMBRACEprot}) are considered of prime importance and are therefore given the highest priority, which is 1. For the added set of aims, all values between the minimum and maximum aspiration values (called aim range) are deemed generally acceptable, so the aims can be adjusted between these two values, while being initialized to the strictest values. They can even be eliminated in patients where the loosest setting cannot be reached as a means of limiting runtime, especially for cases in which an unfavorable applicator implantation leads to a difficult optimization problem. Two of the added aims relate to the normal tissue, which includes any healthy tissue around the target volumes that is not delineated as an OAR. As can be seen in the Supplementary Material Figure 3, the subdivision of ROIs into their `mid-' and `top'- parts is done perpendicularly to the applicator at the depth of the $\mathrm{CTV_{IR}}$ or $\mathrm{CTV_{HR}}$.

\begin{table*}
    \small
    \centering
    \setlength\tabcolsep{2.3pt} %default: 6pt
    \renewcommand{\arraystretch}{1} %default: 1
    \begin{tabular}{|c|c|c|}
         \multicolumn{1}{c}{} & \multicolumn{2}{c}{\textbf{Target volumes}} \\\cline{2-3}
         \multicolumn{1}{c|}{} & $\mathrm{CTV_{HR}}$ & $\mathrm{CTV_{IR}}$ \\\hline
         Aim range {[loose,strict]} & $V_{100\%}>[90,99.9]\%$ & $V_{50\%}>[90,99.9]\%$ \\\hline
         Priority & 2 & 3 \\\hline
    \end{tabular}
    \hspace{.5pt}
    \begin{tabular}{|c|c|c|c|}
         \multicolumn{1}{c}{} & \multicolumn{3}{c}{\textbf{OARs}} \\\cline{2-4}
         \multicolumn{1}{c|}{} & Mid-$\mathrm{CTV_{IR}}$ & Mid-normal-tissue & Top-normal-tissue \\\hline
         Aim range {[strict,loose]} & $V_{100\%}<[25,35]\%$ & $V_{100\%}<[0.1,1.5]\%$ & $V_{100\%}<[0.2,7]\%$ \\\hline
         Priority & 3 & 4 & 4 \\\hline
    \end{tabular}
    \caption{Added set of DVIs, together with their aim ranges and priorities. Aspiration values are defined as percentages of 7 \si{\gray}.}
    \label{tab:addedprot}
    \vspace*{-5mm}
\end{table*}

\subsection{Bi-objective model} \label{subsec:biobjmodel}
The bi-objective model developed for prostate HDR BT intuitively captures the coverage-sparing trade-off by grouping the DVIs into two objectives called the Least Coverage Index (LCI) and the Least Sparing Index (LSI), depending on whether a DVI should be maximized or minimized.
We directly apply the latest version of this bi-objective model \cite{Bouter2019GPU-acceleratedBrachytherapy} to cervical cancer BT, by including the EMBRACE II protocol and the additional set of aims:

\small{
\begin{align}
\begin{split} \label{eq:biobjective}
    \mathrm{LCI}_w(t) &= w_c(\delta_c(D_{90\%}^\mathrm{CTV_{HR}}))
                        + w_c(\delta_c(D_{98\%}^\mathrm{CTV_{HR}}))\\[-0.5mm]
                        &+ w_c(\delta_c(D_{98\%}^\mathrm{GTV_{RES}}))
                        + w_c(\delta_c(D_{98\%}^\mathrm{CTV_{IR}}))\\[-0.5mm]
                        &+ w_c(\delta_{c,\mathrm{adj}}(V_{100\%}^\mathrm{CTV_{HR}}))
                        + w_c(\delta_{c,\mathrm{adj}}(V_{50\%}^\mathrm{CTV_{IR}})),\\[-0.5mm]
    \mathrm{LSI}_w(t) &= w_s(\delta_s(D_{90\%}^\mathrm{CTV_{HR}}))
                        + w_s(\delta_s(D_{2\si{\cubic\centi\metre}}^\mathrm{bladder}))\\[-0.5mm]
                        &+ w_s(\delta_s(D_{2\si{\cubic\centi\metre}}^\mathrm{rectum}))
                        + w_s(\delta_s(D_{\mathrm{point}}^\mathrm{ICRU\ rectovag}))\\[-0.5mm]
                        &+ w_s(\delta_s(D_{2\si{\cubic\centi\metre}}^\mathrm{sigmoid}))
                        + w_s(\delta_s(D_{2\si{\cubic\centi\metre}}^\mathrm{bowel}))\\[-0.5mm]
                        &+ w_s(\delta_{s,\mathrm{adj}}(V_{100\%}^\mathrm{mid-CTV_{IR}}))
                        + w_s(\delta_{s,\mathrm{adj}}(V_{100\%}^\mathrm{mid-normal-tissue}))\\[-0.5mm]
                        &+ w_s(\delta_{s,\mathrm{adj}}(V_{100\%}^\mathrm{top-normal-tissue})),
\end{split}
\end{align}}

\normalsize

where
\small{\begin{gather*}
    \delta_c(D^o_v)=D^o_v-D^{o,\mathrm{aspir}}_v \ \ \mathrm{and} \ \
    \delta_{c,\mathrm{adj}}(V^o_d)=V^o_d-V^{o,\mathrm{aspir}_{\mathrm{adj}}}_d,\\[-0.5mm]
    \delta_s(D^o_v)=D^{o,\mathrm{aspir}}_v-D^o_v \ \ \mathrm{and} \ \
    \delta_{s,\mathrm{adj}}(V^o_d)=V^{o,\mathrm{aspir}_{\mathrm{adj}}}_d-V^o_d,
\end{gather*}}

\normalsize

in which $\mathrm{aspir}_{\mathrm{adj}}$ denotes the aspiration value of the configurable added aim. When optimizing on EMBRACE II aims only, the same model without the $w_s(\delta_{s,\mathrm{adj}}(V^o_d))$ and $w_c(\delta_{c,\mathrm{adj}}(V^o_d))$ terms is used. Weights $w(\delta)$ start at 1 for the least violated DVI and consecutively increase exponentially by a factor of 10; weights are then normalized to numbers between 0 and 1 by dividing by the sum of all weights. When optimizing on each of the objectives, so $\mathrm{LCI}_w(t)$ and $\mathrm{LSI}_w(t)$, the focus thus lies on improving their respective most violated DVI. If $\mathrm{LCI}_w(t)>0$ and $\mathrm{LSI}_w(t)>0$ for a solution, all aims have been reached in that solution.

\subsection{Adaptive configuration of objectives}
We propose an adaptive objective configuration approach for optimization with a tailored version of MO-RV-GOMEA using this bi-objective formulation. It includes the possibility to optimize on two different sets of aims, that are of unequal importance, and of which some of the aspiration values are adjustable. The goal is to ensure that all added aims can be reached, which - since the bi-objective model focuses on the worst-case - allows MO-RV-GOMEA to improve towards all aims, instead of solely concentrating on a potentially non-reachable added aim. The adaptive nature of these aims guarantees that they are still set as strict as possible for each specific patient. The pseudo-code is shown in Algorithm \ref{alg:adaptive} and explained in the following paragraphs.\par
As a reminder, here, a solution $s$ is equivalent to a treatment plan and therefore characterized by DVI values, as well as the corresponding objective values $o_1$ (LCI) and $o_2$ (LSI). The general idea of optimizing with our adaptive objective configuration is to adaptively optimize not only the dwell times, but also properly configure the aspiration values of the added aims. In order to achieve this, MO-RV-GOMEA is run for $g$ generations.
%until convergence is reached; this is done as in MO-RV-GOMEA \cite{Bouter2017TheAlgorithm}, which is summarized in Section \ref{sec:GOMEA}, and includes updating the elitist archive $\mathcal{A}$ with potentially non-dominated solutions of the population $\mathcal{P}$. 
Then, we consider a specific solution $s^*$ in the elitist archive $\mathcal{A}$ by looking at the attained values in objectives $o_1$ and $o_2$ with the goal of determining whether all aims can be met, i.e., find $s^*=\text{arg} \max_{s \in \mathcal{A}}{\{\min{(o_1(s),o_2(s))}\}}$ and see if the adjustable aims are satisfied. If not, the aspiration values of the adjustable DVIs which have not been reached, are adjusted in a stepwise manner and optimization is continued. After every adaptation, objective values and constraints are recomputed for $\mathcal{A}$ and all solutions, and $\mathcal{A}$ are updated. This repeats until all DVI aspiration values of the added aims are achieved. The aspiration values of a DVI in solution $s$ is achieved when $\delta(\text{DVI}_{s}) \geq 0$.\par

\setlength{\textfloatsep}{2pt} %space between float (algorithm) and text
\setlength{\algomargin}{1pt}
\begin{algorithm}
    \small
    \DontPrintSemicolon
    \caption{Adaptive MO-RV-GOMEA-based optimization\\
    \textit{\small{//elitist archive $\mathcal{A}$; population $\mathcal{P}$; solution $s$; objectives $o_i$; \\priorities $p_j$; number of DC points $n_{DC,min}$, $n_{DC,max}$; \\number of generations $g_{n_{DC,min}}$, $g_{n_{DC,max}}$}}}
    \newcommand{\decr}[1]{{\scriptsize {#1}}}
    \label{alg:adaptive}
        \SetKwFunction{CalculateFitnessAndConstraint}{CalculateFitnessAndConstraint}
        \SetKwFunction{RemoveDominatedSolutions}{RemoveDominatedSolutions}
        \SetKwFunction{UpdateElitistArchive}{UpdateElitistArchive}
        \SetKwFunction{SetNumberDCPoints}{SetNumberDCPoints}
        \SetKwFunction{Eliminate}{Eliminate}
        \SetKwFunction{GOMEA}{MO-RV-GOMEA}
        $\text{min\_steps} \gets 4$;
        $\text{adapting} \gets \textsf{true}$\;
        \SetNumberDCPoints{$n_{DC,min}$}\;
        \While{$\textsf{\upshape adapting}$}{
            $\mathcal{A} \gets $ \GOMEA{$g_{n_{DC,min}}$}
             \tcp*{\parbox[t]{.34\linewidth}{\linespread{.5}\selectfont \decr{if first time, then new run, otherwise continued run}}}
            $s^* \gets \text{arg} \max_{s \in \mathcal{A}}{\{\min{(o_1(s),o_2(s))}\}}$\;
            $p_\text{low} \gets \max_{j \in [ 1\ ..\ n_{DVIs} ]}(p_j)$\;
            $\text{adapting} \gets \textsf{false}$\;
            \ForEach {$\text{\upshape DVI}_{s^*}$ $\in$ $\text{\upshape aims}_{\text{\upshape added}}$}{
                \If(\tcp*[f]{\decr{aspiration value not achieved}}){$\delta_\text{adj}(\text{\upshape DVI}_{s^*}) < 0$}{
                    \eIf(\tcp*[f]{\decr{loosest setting not reached}}){$\text{\upshape aspir} != \text{\upshape aspir}_\text{\upshape loose}$}{
                        $\text{aspir} \gets \text{aspir} + \frac{\text{aspir}_\text{loose}-\text{aspir}_\text{strict}}{\text{min\_steps}\cdot(p_\text{low}-p_{\text{DVI}_{s^*}}+1)}$\;
                    }{\Eliminate{$\text{\upshape DVI}_{s^*}$}}
                    $\text{adapting} \gets \textsf{true}$\;
                }
            }
            \If{$\textsf{\upshape adapting}$}{
                \CalculateFitnessAndConstraint{$\mathcal{A}$,$\mathcal{P}$}\;
                \RemoveDominatedSolutions{$\mathcal{A}$}\;
                \UpdateElitistArchive{$\mathcal{A}$,$\mathcal{P}$}
            }
        }
        \SetNumberDCPoints{$n_{DC,max}$}\;
        $\mathcal{A} \gets $ \GOMEA{$g_{n_{DC,max}}$} \tcp*{\decr{new run}}
\end{algorithm}
\setlength{\textfloatsep}{3pt} %space between float (following figures) and text

\setlength{\parskip}{.2pt plus .2pt}

Every adjustable DVI will undergo at least $\text{min\_steps}$. It was set to 4, since this gives a good trade-off between large enough steps to keep computation time sufficiently low for clinical practice, and small enough steps to allow for adequately fine-grained configuration of aspiration values. The step size is taken as the difference between the least strict ($\text{aspir}_\text{loose}$) and the strictest ($\text{aspir}_\text{strict}$) aspiration value defined by the given aim range. Moreover, if a DVI aspiration value has been set to its loosest value, yet still cannot be reached during optimization, then the DVI is eliminated. The different priorities $p$ of the aims are included in terms of the step size of the adjustments made, where smaller steps are attributed to a higher priority. I.e., a higher $p$ value (lower priority) is associated with less steps, so aspiration values are less precisely fine-tuned. The lowest priority $p_\text{low}$ (highest number) of the non-eliminated DVIs gets updated with every continuation of optimization with MO-RV-GOMEA, to adjust the step size after a possible elimination of all lowest priority aims. Note that while performing these adaptations, all DVIs of the EMBRACE II protocol are also actively being optimized on, though their aspiration values are not adjusted.\par

Since multiple adjustment rounds often have to be performed, more than one continuation of optimization with MO-RV-GOMEA is needed, increasing overall runtime. However, during the adjustment phase we are only interested in whether all aims are achievable, not in high-fidelity optimization results. For this reason, we changed the settings of MO-RV-GOMEA during the adjustment phase to achieve a lower runtime. More specifically, we run the optimization with the least possible number of DC points ($n_{DC,min}$), for $g_{n_{DC,min}}$ generations, that still gives a good indication of which aims are satisfied. The last optimization round then is run on a higher number of DC points ($n_{DC,max}$, for $g_{n_{DC,max}}$ generations) as usual. Values for parameters $n_{DC,min}$, $n_{DC,max}$, $g_{n_{DC,min}}$, and $g_{n_{DC,max}}$ are deduced from experiments as described in Section \ref{sec:experiments}.

\vspace*{1pt}
\subsection{Dwell time modulation and catheter contribution restriction}

The intracavitary and possibly interstitial nature of the cervical cancer BT procedure entails the need for a Dwell Time Modulation Restriction (DTMR) and a catheter Contribution Restriction (CR).\par
The DTMR ensures smoothness in neighboring dwell times, which is clinically desirable because it increases treatment plan robustness against displacement uncertainties \cite{Poder2016RobustnessDisplacements}, and can reduce regions of under- and overdose \cite{Mosleh-Shirazi2019InfluenceSagiPlan}. We implemented it by restricting the factor $f$ by which a dwell position with dwell time t can vary from its nearest neighbor - in terms of Euclidean distance - as $f(t)=\frac{2}{5+t}$.
Parameters within this function were set in consultation with medical specialists. If two neighboring dwell times differ by more than $f$, then a constraint value $cons$ is calculated and subtracted as a soft constraint from both objectives:

\vspace*{-4mm}
\small{
\begin{equation}
\begin{gathered}
    cons = \frac{\frac{t}{t_{\mathrm{neighbor}}}-f(t)}{n_{\mathrm{dwells}}},\
    \mathrm{LCI} = \mathrm{LCI} - \alpha\cdot cons,\
    \mathrm{LSI} = \mathrm{LSI} - \alpha\cdot cons,
\end{gathered}
\end{equation}
}
\normalsize
%\vspace*{-3mm}

where $n_{\mathrm{dwells}}$ is the total number of dwell positions and $\alpha=0.01$, which we empirically found to be the lowest $\alpha$ to still lead to the desired dwell time modulation properties.\par

The CR concerns a maximum contribution from the dwell times in the catheters relative to the total dwell time.
%(in applicator and catheters).
Following clinical practice, it was set to 20\% for single catheters, whereas all catheters together can contribute up to 30\%. This restriction is implemented as a hard constraint, rejecting any solutions which violate the given conditions during the optimization.

\section{Experiments} \label{sec:experiments}

Some parameters that we introduced still ought to be tuned. Therefore, this section explains how we determined the smaller number of DC points $n_{DC,min}$ used during the adaptive phase of our approach, and the higher number of DC points $n_{DC,max}$ used in the final run of MO-RV-GOMEA, followed by the determination of their respective maximum number of generations $g_{n_{DC,min}}$ and $g_{n_{DC,max}}$.\par
All experiments are run on an NVIDIA Titan Xp GPU using a parallelized version of MO-RV-GOMEA \cite{Bouter2019GPU-acceleratedBrachytherapy}. Runs are performed on a retrospectively collected dataset of 10 different HDR patient cases, treated between 2017 and 2020 at our university medical center.

\vspace*{-1pt}
\subsection{Dose calculation points}  \label{subsec:experiments:DCpts}

To determine $n_{DC,min}$, two aspects have to be considered. Firstly, optimizing on $n_{DC,min}$ points should give as good an indication of achievable objective values as when a large number of points would be used. Secondly, a non-significant amount of aim adjustments during this optimization should be wrong. To study the first aspect, for each patient case, 30 runs of the original non-adaptive MO-RV-GOMEA are performed, while including the added DVIs, on different numbers of DC points ranging from 500 to 20,000 points per ROI (the default in the clinically introduced MO-RV-GOMEA for prostate BT), resulting in 5,000 to 200,000 points in total. Pareto approximation fronts - of the solutions before reevaluation - are then compared, and the minimum number of DC points $n_{DC,min,obj}$ that gives a sufficiently accurate approximation in terms of obtained LCI and LSI can be determined by visual inspection. Next, to investigate the second aspect, optimization using our adaptive objective configuration is run 30 times, and it is verified whether aim adaptations made per patient per DVI differ significantly between the runs performed on $n_{DC,min,obj}$ DC points per ROI with respect to runs performed on 20,000 DC points per ROI. As such, data for this comparison consists of the number of adjustments per DVI per run for each patient case. Tests are done separately for each patient case and each DVI, where the different runs provide the different data points (each equivalent to the number of adjustments needed in that run). To test for significant difference, paired sample t-tests or Wilcoxon signed rank tests \cite{Wilcoxon1945IndividualMethods} are used, depending on whether Shapiro-Wilk tests \cite{Shapiro1965AnSamples} indicate normally distributed data. Both are Holm-Bonferroni corrected for multiple testing \cite{Holm1979AProcedure}. If no significant difference is found, then $n_{DC,min}=n_{DC,min,obj}$, whereas if there is a significant difference, then $n_{DC,min,obj}$ is increased and the tests are repeated until results are found not to differ.\par
Regarding $n_{DC,max}$, non-adaptive MO-RV-GOMEA with the full set of aims is again run 30 times on each of the 10 patient cases, for different numbers of DC points ranging from $n_{DC,min}$ to 40,000 points per ROI, the latter empirically deemed sufficient while being strictly smaller than the number of DC points used for reevaluation. Subsequently, Pareto approximation fronts before and after reevaluation on 50,000 DC points per ROI (default for the clinical prostate BT implementation) are compared. The lowest number of DC points that results in an accurate calculation of the DVI values, is retained as $n_{DC,max}$. This implies that when there is an acceptable difference \cite{Bouter2019GPU-acceleratedBrachytherapy} between the fallbacks due to two different number of DC points, then $n_{DC,max}$ is the lower number.
%DELETEDBYLEAH
%of DC points.

\vspace*{-1pt}
\subsection{Convergence} \label{subsec:experiments:convergence}

\sloppy
Once $n_{DC,min}$ and $n_{DC,max}$ have been found, the question that is still unanswered is for how many generations MO-RV-GOMEA needs to be run until it has practically converged. This should be determined for $n_{DC,min}$ and $n_{DC,max}$ separately because the fidelity of the fitness function is vastly different, leading to distinct resource requirements. As we are at this stage only interested in single optimization rounds (e.g., between two adaptations), the original non-adaptive MO-RV-GOMEA is run.\par
Generally, to decide at which generation convergence is approximately reached, we compare the best obtained LCI value during every generation. Since dwell times are initialized between 0 and 2 \si{\second} - which was empirically determined -, the optimization starts with a low coverage, so the LCI is strictly increased over time and thereby representative for convergence. We take as a reference for what is obtainable the maximum LCI after 20 \si{\minute} ($\mathrm{LCI}_{20\si{\minute}}$). This is considered a more than appropriate time, since, firstly, visual inspection of the Pareto approximation plots after 20 \si{\minute} indicate convergence, and, secondly, the algorithm has been shown to converge after 3 \si{\minute} on 100,000 DC points (in total, for all ROIs) in the prostate case on the same hardware \cite{Bouter2019GPU-acceleratedBrachytherapy}, and 25,000 DC points in total are used for the adaptive rounds in this work for the cervix.\par
Thus, the algorithm is considered to have converged at the first generation $g$, for which 99\% of $\mathrm{LCI}_{20\si{\minute}}$ is reached (taken as a difference with respect to $\mathrm{LCI}_1$: the maximum LCI in generation 1), and for which the change in maximum LCI for the subsequent 20 generations is below $10^{-4}$, i.e, the following two conditions hold:

%\small
\vspace*{-3mm}
\begin{equation}\label{eq:convcond1}
%    \frac{\mathrm{LCI}_{g}-\mathrm{LCI}_1}{\mathrm{LCI}_{20\si{\minute}}-\mathrm{LCI}_1} > 99\% 
    \left(\mathrm{LCI}_{g}-\mathrm{LCI}_1\right)\ /\ \left(\mathrm{LCI}_{20\si{\minute}}-\mathrm{LCI}_1\right) > 99\%,
\end{equation}
\begin{equation}\label{eq:convcond2}
    \mathrm{number\_generations\_after\_}g(\Delta\mathrm{LCI}<10^{-4}) \geq 20.
\end{equation}
%\normalsize

\noindent
The first generation $g$ when (\ref{eq:convcond1}) and (\ref{eq:convcond2}) are met, is found separately for 5 runs of 20 \si{\minute} for every patient case. Then, the maximum found $g$ is retained. This is done for optimizations on $n_{DC,min}$ as well as on $n_{DC,max}$, to obtain respectively $g_{n_{DC,min}}$ and $g_{n_{DC,max}}$.
\fussy

\vspace*{-1pt}
\subsection{Evaluation} \label{subsec:experiments:evaluation}

The main goal of the proposed adaptive objective configuration is to generate treatment plans that are clinically better than the ones achieved through standard optimization on only the clinical aims as defined in the EMBRACE II protocol. Therefore, it is essential to compare treatment plans generated by the two approaches. This can be subdivided into three different aspects, given 
below. To allow for a fair comparison with regards to precision of the final result and potential fallback during reevaluation, the optimization solely based on the EMBRACE II protocol is run on $20,000$ DC points per ROI, which corresponds to the found $n_{DC,max}$ (see section \ref{subsec:results:DCpts}). Both approaches are reevaluated on $50,000$ DC points per ROI, and are run 30 times for each patient case.\par

First, the adaptive approach should ideally perform as well as the original approach in terms of achieved EMBRACE II aims. We thus, for each approach, contrast the number of runs (out of 30 in total) for which all of the EMBRACE II aims were satisfied in at least one treatment plan in the Pareto approximation front.\par

Second, the most important undesired properties of the plans found by optimizing on only the aims of the EMBRACE II protocol are directly captured by the values associated with the added DVIs. Hence, we compare the achieved values for the added DVIs for each method. As the main clinical interest lies in the plans that satisfy the EMBRACE II aims, only the DVIs of the plans that do so are considered, while all other plans are discarded. The number of concerned runs can be directly taken into account since it is presented alongside (see first evaluation point), and the mean number of plans (in which the EMBRACE II aims were satisfied) per run is also presented for the two approaches.
\par

Third, it is essential to compare the dose distributions resulting from the two methods.
One plan from the first run of every patient case is selected: from the whole generated set of plans, plan $s^*=\text{arg} \max_{s \in \mathcal{A}}{\{\min{(o_1(s),o_2(s))}\}}$ is considered (regarded best in both objectives). Then, dose distributions and DVI values associated with plan $s^*$ from each patient case are compared in clinically approved software (Oncentra Brachy (Elekta, Veenendaal, The Netherlands)). This is done by a team of medical specialists 
(a radiation oncologist, BT medical physicist, and BT technologist), 
who subsequently choose the preferred plan - which they would like to immediately use or take as a starting point for small additional manual adaptations - between the two optimization approaches.

\vspace*{8pt}
\section{Results and Discussion}
We first describe the results from the parameter tuning experiments pertaining to Sections \ref{subsec:experiments:DCpts} and \ref{subsec:experiments:convergence}. Then, results from the three evaluation aspects in Section \ref{subsec:experiments:evaluation} are presented.

\vspace*{8pt}
\subsection{Dose calculation points}  \label{subsec:results:DCpts}

Results of optimizations on 500 to 20,000 DC points per ROI for the non-adaptive MO-RV-GOMEA before reevaluation are shown in Figure \ref{fig:DCmin-reach} for two patient cases, others are in Supplementary Material Figure 4. By visual inspection, it can be observed that 500 and 1,000 DC points are not enough to accurately calculate the DVI values, since often an under- or overestimation of LCI and LSI values can be observed. Using 2,500 DC points onwards provides good estimations of reachable LCI and LSI values, so $n_{DC,min,obj}=$ 2,500. The apparent discrete nature of the front is due to imprecisions occurring when calculating the objective values, and does not impact the reachable LCI and LSI values, which is the only goal at this stage.\par

\begin{figure}[h]
  \vspace*{3pt}
  \centering
  \includegraphics[width=0.92\linewidth]{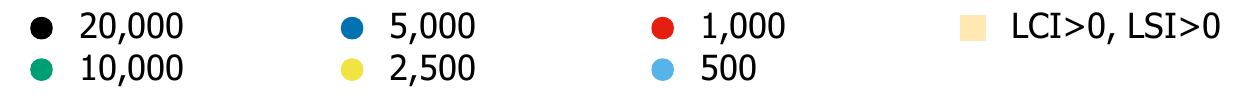}
  \vspace{4pt}
  \includegraphics[width=\linewidth]{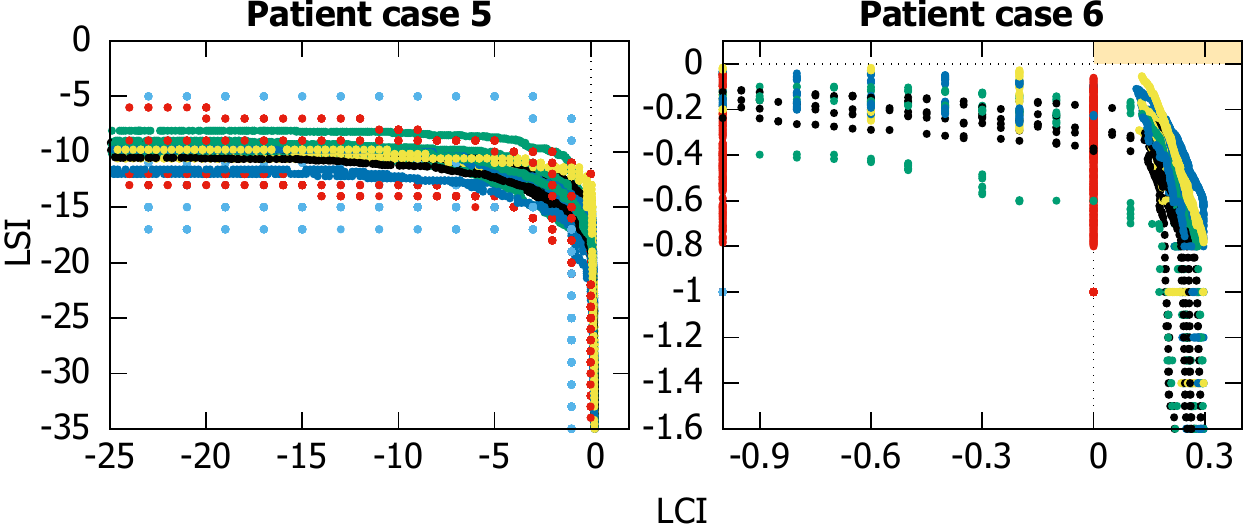}
  \vspace*{-10pt}
  \caption{Pareto approximation fronts before reevaluation for optimizations on different numbers of DC points per ROI. Each plot shows the superposition of five runs per number of DC points. All plans with $\text{LCI}>0$ and $\text{LSI}>0$ (shaded beige) satisfy all aims.
  The axis ranges differ between the two cases.}
  \label{fig:DCmin-reach}
  \vspace*{12pt}
\end{figure}

Taking $n_{DC,min,obj}=$ 2,500 for the adaptive rounds of the new optimization approach, data retrieved on number of adjustments made per aim (for each run and each patient) are non-normally distributed in 97\% of the cases (Shapiro-Wilk tests, $\mathrm{p}<0.039$).

\setlength{\parskip}{.2pt plus .2pt}

We therefore use the Wilcoxon signed rank test to evaluate whether differences in the number of made adjustments between $n_{DC,min,obj}=$ 2,500 and 20,000 DC points are significant. The null hypothesis is that there is no difference between the two groups, and we use a significance level of $\alpha=0.05$. Since $n>20$ (not counting ties), an estimation based on the normal approximation can be made. As ties are frequently present in the data (the number of adjustments is often the same for different runs), we first correct the p-values for ties. Moreover, we use a continuous approximation for a discrete distribution, so we further correct the p-values for continuity, which is done by subtracting a continuity correction factor of $\frac{0.5}{\sigma}$ from the z-score \cite{Ross2004DistributionsStatistics}, where $\sigma$ is the standard deviation. Then, $\alpha$ values are Holm-Bonferroni corrected, because multiple significance tests are done simultaneously. The found p-values, as well as $\alpha$ values for each of the adjustable aims and patient cases are presented in the Supplementary Material Table 1.
%Table \ref{tab:pvalues}. 
We reject the null hypothesis if $\mathrm{p}<\alpha$, in which case there is a significant difference between the groups of different number of DC points, whereas if $\mathrm{p}>\alpha$, then there is no significant difference.
Since $\mathrm{p}>\alpha$ for all patient cases, we find that there is no significant difference between the number of adjustments made for each aim when using 2,500 versus 20,000 DC points. We conclude that $n_{DC,min,obj}=n_{DC,min}=$ 2,500.\par

Regarding the number of DC points $n_{DC,max}$ for the last optimization round, the Pareto approximation fronts before and after reevaluation are shown in Figure \ref{fig:DCmax} (and in Supplementary Material Figure 5). The six shown optimization runs were performed on a number of DC points per ROI between 2,500 and 40,000, and reevaluation was done on 50,000 DC points per ROI. As expected, the fallback of the plans in achieved LCI and LSI values decreases with increasing number of DC points, because a higher number of DC points implies a more precise calculation of the objective values. Since we usually observe an acceptable difference between the fallbacks for 20,000 versus 30,000 (and 40,000) DC points, 20,000 is considered enough for accurate calculation of LCI and LSI values. We thus set $n_{DC,max}=$ 20,000. This is in line with results found for prostate cancer BT~\cite{Bouter2019GPU-acceleratedBrachytherapy}. Of note is that while there is a considerable fallback with 2,500 DC points, the approximation front before evaluation (green points) has similar LCI and LSI values as when more DC points are used, confirming our findings for $n_{DC,min}$.
\setlength{\parskip}{0pt}

\begin{figure}[h]
  \centering
  \includegraphics[width=\linewidth]{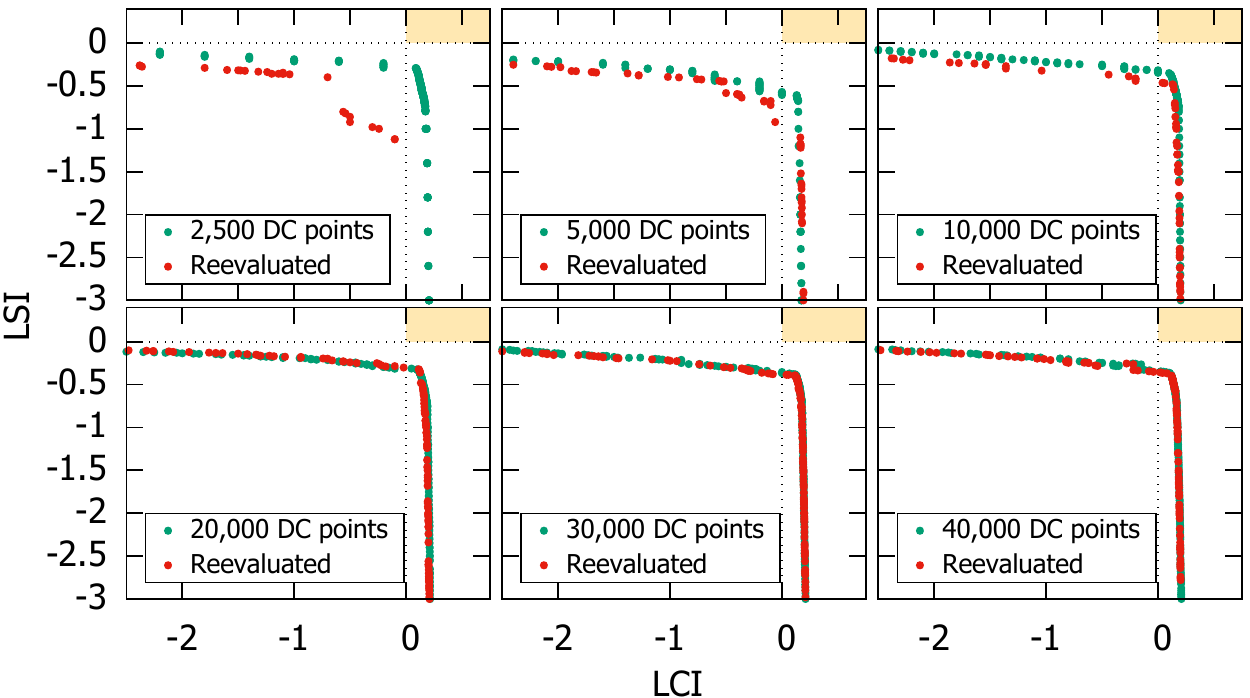}
  \vspace*{-18pt}
  \caption{Comparison of Pareto approximation fronts (of one run) before and after reevaluation on 50,000 DC points per ROI on patient case 1. Optimization was done on 2,500 to 40,000 DC points per ROI. All plans with $\text{LCI}>0$ and $\text{LSI}>0$ (shaded beige) satisfy all aims (EMBRACE II and added).}
  \label{fig:DCmax}
  \vspace*{5pt}
\end{figure}

\subsection{Convergence} \label{subsec:results:convergence}

Based on the conditions given in Section \ref{subsec:experiments:convergence}, the maximum $g$ values for the non-adaptive MO-RV-GOMEA for cervical cancer on $n_{DC,min}=$ 2,500 DC points per ROI are found for 10 patient cases. These maxima have a mean of 267 and ranged from 180 to 346, which was rounded to $g_{n_{DC,min}}=350$.\par
The same experiments for $n_{DC,max}=20,000$ DC points per ROI lead to maxima over all patient cases with a mean of 307 generations, ranging from 203 to 489, which we rounded to $g_{n_{DC,max}}=490$.

{\renewcommand{\arraystretch}{0.9}
\begin{table*}[h!]
    \newcolumntype{?}{!{\vrule width 1pt}}
    \makeatletter
        \def\hlinewd#1{%
        \noalign{\ifnum0=`}\fi\hrule \@height #1 \futurelet
        \reserved@a\@xhline}
        \makeatother
    \small
    \centering
    \newcommand{\nsig}{\cellcolor{green!10}}
    \newcommand{\sig}{\cellcolor{red!10}}
    \begin{tabular}{c?c?ccccccccccc}
        \multicolumn{2}{c?}{Patient case} & 1 & 2 & 3 & 4 & 5 & 6 & 7 & 8 & 9 & 10 \\\hlinewd{1pt}
        E satisfied & F & 100\% & 97\% & 97\% & 93\% & 0\% & 87\% & 87\% & 0\% & 100\% & 100\% \\
           (\% of runs) & E & 100\% & 100\% & 100\% & 100\% & 33\% & 100\% & 100\% & 0\% & 100\% & 100\% \\\hlinewd{1pt}
        \multirow{2}{*}{$V_{50\%}^{\mathrm{CTV_{IR}}}$} & F & 99.9 $\pm$ 0.2 & \: 99.2 $\pm$ 0.3 & 98.7 $\pm$ 0.4 & 99.6 $\pm$ 0.3 & \textit{n.a.} & 98.8 $\pm$ 0.5 & \: 98.5 $\pm$ 0.3 & \textit{n.a.} & \: 99.2 $\pm$ 0.3 & 100.0 $\pm$ 0.0 \\
           & E & 99.5 $\pm$ 0.4 & 100.0 $\pm$ 0.0 & 99.7 $\pm$ 0.1 & 99.9 $\pm$ 0.1 & 98.4 $\pm$ 0.3 & 99.9 $\pm$ 0.5 & 100.0 $\pm$ 0.0 & \textit{n.a.} & 100.0 $\pm$ 0.1 & 100.0 $\pm$ 0.0 \\\hline
        \multirow{2}{*}{$V_{100\%}^{\mathrm{mid-CTV_{IR}}}$} & F & 24.1 $\pm$ 1.7 & 25.7 $\pm$ 2.0 & 24.0 $\pm$ 2.3 & 23.3 $\pm$ 2.4 & \textit{n.a.} & 24.8 $\pm$ 1.8 & 25.3 $\pm$ 1.7 & \textit{n.a.} & 27.9 $\pm$ 2.3 & 27.0 $\pm$ 2.2 \\
           & E & 28.2 $\pm$ 3.0 & 35.0 $\pm$ 2.8 & 39.0 $\pm$ 3.8 & 25.5 $\pm$ 3.8 & 35.2 $\pm$ 1.7 & 38.8 $\pm$ 3.9 & 40.2 $\pm$ 2.6 & \textit{n.a.} & 39.5 $\pm$ 3.3 & 33.4 $\pm$ 3.6 \\\hline
        \multirow{2}{*}{$V_{100\%}^{\mathrm{CTV_{HR}}}$} & F & 97.9 $\pm$ 0.4 & 98.4 $\pm$ 0.4 & 97.6 $\pm$ 0.7 & 98.2 $\pm$ 0.6 & \textit{n.a.} & 98.0 $\pm$ 0.7 & 98.5 $\pm$ 0.6 & \textit{n.a.} & 98.7 $\pm$ 0.5 & 98.5 $\pm$ 0.5 \\
           & E & 97.6 $\pm$ 0.6 & 98.2 $\pm$ 0.6 & 97.8 $\pm$ 0.7 & 98.2 $\pm$ 0.7 & 96.1 $\pm$ 0.4 & 97.2 $\pm$ 0.7 & 98.6 $\pm$ 0.7 & \textit{n.a.} & 99.2 $\pm$ 0.5 & 98.8 $\pm$ 0.5 \\\hline
        \multirow{2}{*}{$V_{100\%}^{\mathrm{mid-NT}}$} & F & 0.2 $\pm$ 0.1 & 0.4 $\pm$ 0.1 & 0.2 $\pm$ 0.1 & 0.1 $\pm$ 0.1 & \textit{n.a.} & 0.3 $\pm$ 0.2 & 0.4 $\pm$ 0.2 & \textit{n.a.} & 0.7 $\pm$ 0.1 & 0.3 $\pm$ 0.1 \\
           & E & 1.4 $\pm$ 0.5 & 0.8 $\pm$ 0.1 & 1.5 $\pm$ 0.9 & 0.6 $\pm$ 0.3 & 1.8 $\pm$ 0.1 & 2.4 $\pm$ 1.0 & 2.3 $\pm$ 0.3 & \textit{n.a.} & 1.2 $\pm$ 0.2 & 0.7 $\pm$ 0.2 \\\hline
        \multirow{2}{*}{$V_{100\%}^{\mathrm{top-NT}}$} & F & 0.0 $\pm$ 0.0 & 0.0 $\pm$ 0.1 & 0.1 $\pm$ 0.2 & 0.1 $\pm$ 0.5 & \textit{n.a.} & 0.0 $\pm$ 0.0 & 0.6 $\pm$ 0.9 & \textit{n.a.} & 0.1 $\pm$ 0.1 & 0.0 $\pm$ 0.1 \\
           & E & 0.0 $\pm$ 0.0 & 1.4 $\pm$ 0.9 & 3.4 $\pm$ 1.2 & 1.6 $\pm$ 0.6 & 1.3 $\pm$ 0.7 & 0.8 $\pm$ 0.4 & 6.3 $\pm$ 1.3 & \textit{n.a.} & 1.2 $\pm$ 0.3 & 0.4 $\pm$ 0.2 \\
    \end{tabular}
    \caption{Adaptive optimization on the full (EMBRACE II and added) set of DVIs (F) versus optimization on EMBRACE II DVIs (E) only. First row: percentage of 30 runs for which all EMBRACE II aims were satisfied. Subsequent rows: median$\pm$st.dev. values over 30 runs for each adjustable DVI. Abbreviations: \textit{n.a.}: EMBRACE II aims were never all satisfied; NT: normal tissue.}
    \label{tab:evaluation}
\end{table*}
}

\begin{figure*}[h!]
\vspace{-14pt}
  \centering
  \begin{minipage}{.85\linewidth}
      \includegraphics[width=.97\linewidth,right]{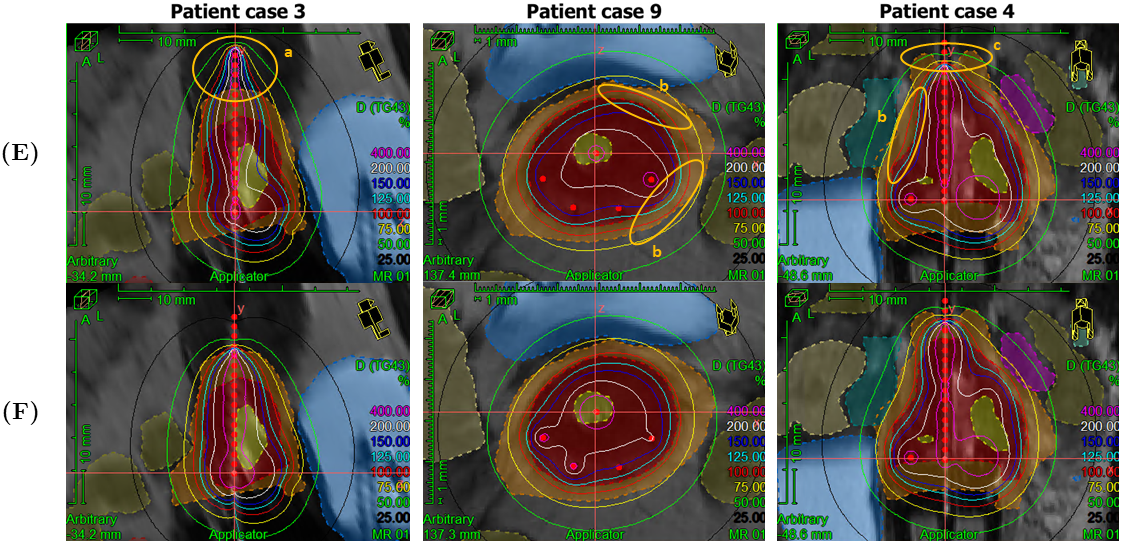}
  \end{minipage}
  \hspace{6pt}
  \begin{minipage}{.1\linewidth}
      \includegraphics[width=0.82\linewidth,left]{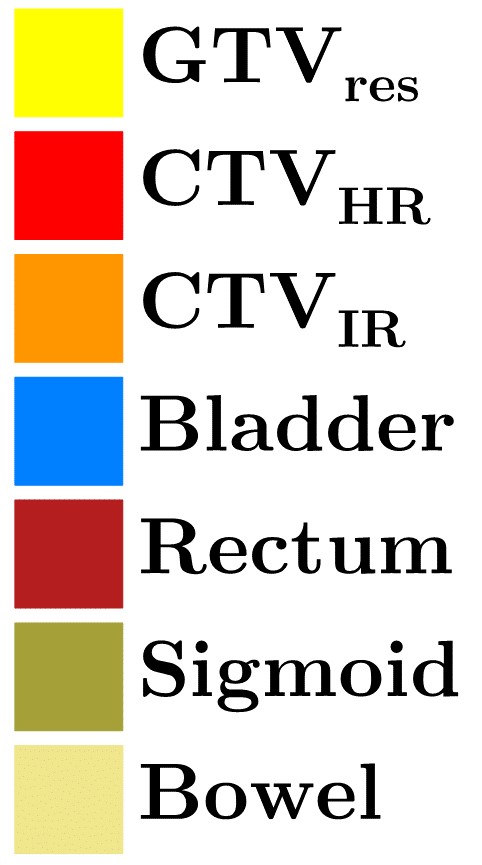}
  \end{minipage}
\vspace{-6pt}
  \caption{Dose distributions of plans: EMBRACE II optimized (top) versus adaptively optimized on the full set of DVIs (bottom). Bottom plans were preferred over top plans for the three shown cases. Undesirable properties are circled in orange: (a) high doses to normal tissue; (b) high doses outside the $\mathrm{CTV_{IR}}$; (c) low doses to the $\mathrm{CTV_{IR}}$. Red points denote dwell positions. Dotted lines and color-shaded areas delineate target volumes and OARs, while solid colored lines are isodose lines, encompassing areas that receive doses of 25\% up to 400\% of 7 \si{\gray}. 
  %View directions are given in the top right corner of each image. 
  Visualized with Oncentra Brachy (Elekta, Veenendaal, The Netherlands).}
  \label{fig:distributions}
\vspace{-7pt}
\end{figure*}

\vspace*{-2pt}
\subsection{Evaluation} \label{subsec:results:evaluation}

For each approach, the number of runs that lead to all EMBRACE II DVIs being satisfied for at least one treatment plan in the approximation front, and the DVI values of the added aims associated with the plans for which the EMBRACE II aims were achieved, are summarized in Table \ref{tab:evaluation}. Results for each of the aspects are described below. We abbreviate the two approaches as follows:

\begin{itemize}[topsep=0.5mm,leftmargin=5mm]
    \item (E): optimization on EMBRACE II aims, with the original non-adaptive MO-RV-GOMEA,
    \item (F): optimization on full set of aims (EMBRACE II and added ones), with the adaptive objective configuration approach.
\end{itemize}

\noindent
Firstly, the number of runs in which all EMBRACE II DVIs were satisfied is given in percentages (from 30 runs in total) in the top row in Table \ref{tab:evaluation}. It indicates that (E) leads to 8/10 patient cases for which 100\% of the runs reached the EMBRACE II aims. In contrast, (F) leads to only 3/10 patient cases satisfying the EMBRACE II DVIs in 100\% of the runs. However, except for cases 5 and 8, in the runs that did not achieve all EMBRACE II aims, either the coverage or the sparing aims were achieved, and neither LCI nor LSI were inferior to $-0.05$.
%Per patient case, the percentages were reduced by maximum 30\% of runs (case 7). 
This nonetheless does indicate that there is a compromise in reaching the EMBRACE II aims when introducing new aims with the adaptive optimization approach. A potential way of improving upon this is not to stop the adaptive rounds when all added aims are satisfied (stop condition 1), but instead, after that, use a second stop condition of having reached the EMBRACE II aims. For difficult patients, for which the EMBRACE II aims cannot be reached, the previous set of solutions (obtained at satisfying stop condition 1) would then be presented. This would ensure that the same amount of runs as in (E) reach the EMBRACE II aims, though the runtime would considerably increase.\par

\setlength{\parskip}{.2pt plus .2pt} %otherwise no stretchable vertical space so end of columns do not match up

Secondly, the DVI values of the added aims are presented in Table \ref{tab:evaluation}, which correspond to medians and standard deviations over the 30 runs, and are given in percentages of 7 \si{\gray} as 
in the protocol (see Table \ref{tab:addedprot}). For the added DVIs which have a minimization aim associated with them, i.e., $V_{100\%}^{\mathrm{mid-CTV_{IR}}}$, $V_{100\%}^{\mathrm{mid-NT}}$ and $V_{100\%}^{\mathrm{top-NT}}$, attained medians were lower in (F) than in (E), as is expected, since (F) optimizes for them to be minimized. Differences are especially considerable in patient cases 3 and 7. However, for the added DVIs which have a maximization aim ($V_{50\%}^{\mathrm{CTV_{IR}}}$ and $V_{100\%}^{\mathrm{CTV_{HR}}}$), one can see that obtained median values do not differ much between (E) and (F), and some are even higher for (E) while they were not optimized on in that approach. This nonetheless does not imply that these aims could have been omitted in the optimization, since the added maximization and minimization DVIs describe conflicting objectives. The normal tissue is situated directly around the $\mathrm{CTV_{IR}}$, so satisfying both a minimum dose to the $\mathrm{CTV_{IR}}$ and a maximum dose to the normal tissue DVIs is desirable, but often impossible to reach. Of note is also that a mean of 19.9 plans per run were found to reach the EMBRACE II aims for (E), whereas the mean for (F) was 109.9. This should however not be seen as a drawback since it was to be expected that the added set of aims would be in conflict with the EMBRACE II aims. Most importantly, a diverse set of treatment plans to choose from that trade off the additional aims is now found, which gives key insights into what is possible for the specific patient at hand.\par

Thirdly, when presented with the dose distributions and DVI values of one plan from each approach, medical specialists selected the plan resulting from (F) over the plan resulting from (E) in 8/10 patient cases. 
In 2/10 cases, (E) was preferred over (F) after some discussion, criticisms of the plans resulting from (F) having a dose distribution which was too erratic (i.e., non-smooth isodose lines when scrolling through different slices) and slightly too much dose to the bladder and sigmoid. An area which was generally criticized was around the base of the applicator, which is as of now not included directly in the optimization. Future research would be necessary for a minimization of that dose, and for a prioritization of applicator dwell positions over catheter dwell positions. An impression of the dose distributions is provided in Figure \ref{fig:distributions} for three different patient cases. 
One can see that the marked unwanted properties in the plans obtained by (E) are clearly diminished in the plans resulting from (F). It is worth noting that not all differences between the shown dose distributions are due to the distinct optimization methods, since MO-RV-GOMEA is of stochastic nature. Moreover, only one plan from the approximation set is shown. \par

Finally, an essential part of the evaluation of an optimization approach for BT
is its runtime. The proposed adaptive approach had a maximum runtime of 11.4 \si{\minute} (over 30 runs and 10 patient cases, minimum 2.9 \si{\minute}, median 5.3 \si{\minute}, standard deviation 1.6 \si{\minute}), excluding initialization and DC points sampling times. Even though this might not surpass the time used for the current manual treatment planning, we acknowledge that it exceeds a desirable amount of time for automatic planning, especially if manual adjustments would still be needed on the chosen automatically generated plan. Further research is being done to speed up the procedure. For instance, runtime could be decreased by setting a different minimum number of DC points per specific ROI, as not every ROI might need the same number of DC points, since they vary in size and closeness to the dwell positions and thus planned dose.\par

\vspace*{-2pt}
\section{Conclusions}

To expand the success in generating treatment plans for prostate brachytherapy, we tailored the bi-objective MO-RV-GOMEA based approach to the case of cervical cancer brachytherapy. In contrast to prostate brachytherapy, directly optimizing on DVIs from a base clinical protocol
did not lead to clinically desirable treatment plans. Additional aims needed to be considered that have other priorities and adjustable aspiration values. We therefore designed an alternative MO-RV-GOMEA-based optimization approach capable of using an added set of adaptively configurable DVIs. We configured and evaluated our new approach on 10 patient cases. After incorporating key added aims identified in consultation with clinical experts, overall, unwanted properties in found plans as judged both in terms of DVI values and in the dose distributions were successfully reduced to clinical satisfaction. Nonetheless, most of the resulting plans can still be improved, and the runtime of median 5.3 \si{\minute}, especially for difficult patients, exceeds desirable times. Our main contribution, however, is the development of an optimization approach that supports adaptively adjustable aims for each patient, which can effortlessly be added or removed. To ultimately arrive here in (nearly) 100\% of all cases, iterations of identifying and adding new aims and re-running optimization on cohorts of patients is required. In light of a clinical protocol itself not sufficing, having this ability is key to achieve clinically desirable plans.

%-----------------------------------------------------------------------------
%%
%% The acknowledgments section is defined using the "acks" environment
%% (and NOT an unnumbered section). This ensures the proper
%% identification of the section in the article metadata, and the
%% consistent spelling of the heading.
\vspace*{-2pt}
\begin{acks}

This work was funded by the Dutch Cancer Society (KWF Kankerbestrijding; Project N. 12183) and by Elekta.

\end{acks}

%%
%% The next two lines define the bibliography style to be used, and
%% the bibliography file.
\bibliographystyle{ACM-Reference-Format}
\bibliography{references, manual}

%%
%% If your work has an appendix, this is the place to put it.

%\appendix
%\section{}

\end{document}